\documentclass[sigconf]{acmart}

\usepackage{multirow,tabularx}
\usepackage{caption}
\usepackage{subcaption}
  \usepackage{placeins}

\AtBeginDocument{%
  \providecommand\BibTeX{{%
    \normalfont B\kern-0.5em{\scshape i\kern-0.25em b}\kern-0.8em\TeX}}}

\setcopyright{acmcopyright}
\copyrightyear{2023}
\acmYear{2023}
\acmDOI{XXXXXXX.XXXXXXX}

\acmPrice{15.00}
\acmISBN{978-1-4503-XXXX-X/18/06}

\begin{document}

\title{WildGEN: Long-horizon Trajectory Generation for Wildlife}

\author{Ali Al-Lawati}
\email{aha112@psu.edu}
\orcid{1234-5678-9012}
\affiliation{%
  \institution{Penn State}
  \streetaddress{P.O. Box 1212}
  \city{State College}
  \state{Pennsylvania}
  \country{USA}
  \postcode{16802}
}

\author{Elsayed Eshra}
\email{eme5375@psu.edu}

\affiliation{%
  \institution{Penn State}
  \streetaddress{P.O. Box 1212}
  \city{State College}
  \state{Pennsylvania}
  \country{USA}
  \postcode{16802}
}

\author{Prasenjit Mitra}
\email{pum10@psu.edu}
\affiliation{%
  \institution{Penn State}
  \streetaddress{P.O. Box 1212}
  \city{State College}
  \state{Pennsylvania}
  \country{USA}
  \postcode{16802}
}

\renewcommand{\shortauthors}{}

\begin{abstract}
  Trajectory generation is an important concern in pedestrian, vehicle, and wildlife movement studies. Generated trajectories help enrich the training corpus in relation to deep learning applications, and may be used to facilitate simulation tasks. This is especially significant in the wildlife domain, where the cost of obtaining additional real data can be prohibitively expensive, time-consuming, and bear ethical considerations. In this paper, we introduce WildGEN: a conceptual framework that addresses this challenge by employing a Variational Auto-encoders (VAEs) based method for the acquisition of movement characteristics exhibited by \textit{wild geese} over a long horizon using a sparse set of truth samples. A subsequent post-processing step of the generated trajectories is performed based on smoothing filters to reduce excessive wandering. Our evaluation is conducted through visual inspection and the computation of the Hausdorff distance between the generated and real trajectories. In addition, we utilize the Pearson Correlation Coefficient as a way to measure how \begin{it}realistic\end{it} the trajectories are based on the similarity of clusters evaluated on the generated and real trajectories.
\end{abstract}

\begin{comment}
\begin{CCSXML}
<ccs2012>
 <concept>
  <concept_id>00000000.0000000.0000000</concept_id>
  <concept_desc>Do Not Use This Code, Generate the Correct Terms for Your Paper</concept_desc>
  <concept_significance>500</concept_significance>
 </concept>
 <concept>
  <concept_id>00000000.00000000.00000000</concept_id>
  <concept_desc>Do Not Use This Code, Generate the Correct Terms for Your Paper</concept_desc>
  <concept_significance>300</concept_significance>
 </concept>
 <concept>
  <concept_id>00000000.00000000.00000000</concept_id>
  <concept_desc>Do Not Use This Code, Generate the Correct Terms for Your Paper</concept_desc>
  <concept_significance>100</concept_significance>
 </concept>
 <concept>
  <concept_id>00000000.00000000.00000000</concept_id>
  <concept_desc>Do Not Use This Code, Generate the Correct Terms for Your Paper</concept_desc>
  <concept_significance>100</concept_significance>
 </concept>
</ccs2012>
\end{CCSXML}

\ccsdesc[500]{Do Not Use This Code~Generate the Correct Terms for Your Paper}
\ccsdesc[300]{Do Not Use This Code~Generate the Correct Terms for Your Paper}
\ccsdesc{Do Not Use This Code~Generate the Correct Terms for Your Paper}
\ccsdesc[100]{Do Not Use This Code~Generate the Correct Terms for Your Paper}
\end{comment}

%%\received{20 February 2007}
%%\received[revised]{12 March 2009}
%%\received[accepted]{5 June 2009}

\maketitle

\section{Introduction}
Animals in the wild move for a variety of reasons, including foraging, exploration, or migration \cite{venter_intrinsic_2015}. A better understanding of the drivers of wildlife movement can have tremendous benefits in environmental planning, preservation efforts, and anti-poaching. It contributes to mitigating conflict between wildlife and human settlements \cite{main_human_2020}.

Scientists in wildlife studies use datasets of animal trajectories collected, typically, using telemetry devices, such as GPS collars, to capture the animals' exact location over time. However, the limitations of using these devices arise from the complications and dangers associated with attaching the devices to the animals, including constraints on device size and weight depending on the animal, potential failures, and battery life. 

As a result, considerable work in movement ecology attempts to generate wildlife trajectories based on concepts such as Correlated Random Walk (CRW) \cite{renshaw1981correlated} and Brownian Motion \cite{horne2007analyzing}. However, some of these methods are limited in their effectiveness to land-based animals. Furthermore, in their basic form, these methods can only model the wandering effect of animals within a foraging area \cite{bailey2006quantitative}. Methods such as Brownian Bridges \cite{technitis_b_2015} extend these movement models using a beginning and endpoint, and interpolate the in-between points based on the maximum speed and time budget of the animal over the expected duration of the trajectory. However, these methods fail to take existing measured trajectories as input in the process of trajectory generation. Instead, they rely on random variables to induce movement, and thereby achieve limited success. 

On the other hand, pedestrian and vehicle trajectory generation problems are constrained by the road and pedestrian networks which are not applicable in the wildlife domain. Likewise, trajectory generation in the UAV literature is irrelevant as it is more concerned with optimization, and minimizing flying time through the target waypoints \cite{singh2001trajectory}.
 \begin{figure}
\includegraphics[width=0.5\textwidth]{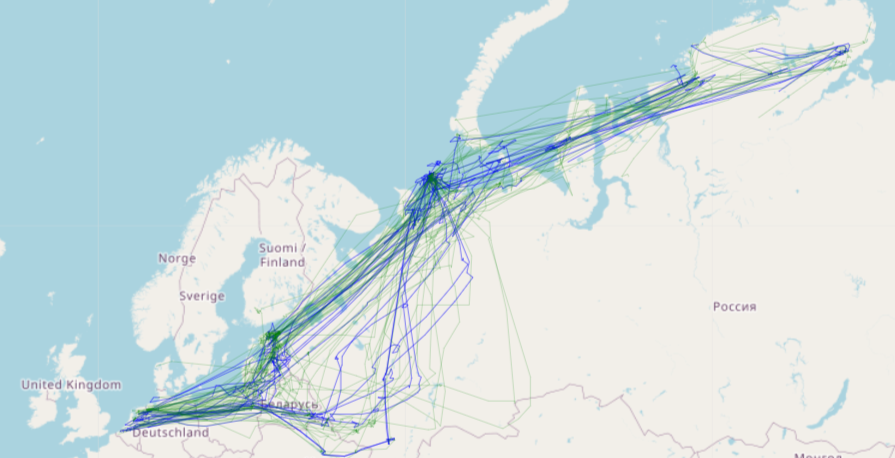} 
\centering
\caption{Real (green) vs Generated (blue) Trajectories}
%%\begin{tabular}{r@{: }l r@{: }l}
%%Green & Real Trajectories & Red & Generated Trajectories 
%%\end{tabular}
\label{fig:1}
\end{figure}

In this paper, we present a novel framework we call WildGEN that attempts to generate trajectories by learning the statistical properties of a sparse set of known trajectories. The framework, WildGEN, is based on proven deep learning-based Variational Auto-encoders (VAEs) which have successfully been applied to similar problems \cite{trajvae}. WildGEN applies a post-processing filter on each trajectory path using a Minimum Bounding Region (MBR) based constraint. Furthermore, in order to reduce the \begin{it}excessive wandering effect\end{it}, a smoothing filter is applied. 

In the next section, we discuss some related work, followed by a brief motivation of the methods we used to develop our Framework. Section 4 presents the experimental settings, experimental results, and an ablation study that demonstrates the contribution of each component to the overall results. We conclude and discuss future work in Section 5.
%%[why don't we generate trajectories between known clusters and concatenate them] Finally, a GAN is used to discriminate the resulting trajectory based on the real dataset and exclude improbable results.

%[idea can use Savitzky-Golay for smooting]
%[input/output structure - model structure (number of layers/number of neurons/activation %functions/optimizer) - loss function -  ]

The contributions of this work can be summarized as follows:
\begin{itemize}
    \item We propose a framework that generates realistic long-horizon wildlife trajectories based on a sparse set of real trajectories.
    \item We experimentally test our framework and we validate the empirical results using Hausdorff distance and the Pearson Correlation Coefficient metrics to measure path and cluster similarities, respectively between the real and generated trajectory sets.
    \item We benchmark our method against Levy Walk/Flight \cite{zaburdaev2015levy}, which is state-of-the-art in long-horizon trajectory generation in the wildlife domain. We also evaluate our methods over kriging \cite{iglesias2018spatio}, using a Heteroscedastic Gaussian Process Regression (GPR) of multiple trajectories on the same time horizon.
\end{itemize}
{Figure \ref{fig:1} visually demonstrates the potential of this approach. We further evaluate our approach using well-established trajectory similarity measures.}

\subsection{Problem Statement}

Given a set of $n$ real trajectories, each consisting of $m$ $(x, y)$ points, the objective is to produce a set of synthetic trajectories that maximize the following:

\begin{enumerate}
    \item (Path Similarity) How similar are the points of the synthetic trajectories to the real trajectories?
    \item (Cluster Similarity) How well do the synthetic trajectories align with the real trajectories in terms of areas of concentration?
\end{enumerate}
Formally, given $n$ real trajectories, denoted as $$T_1, T_2, \ldots, T_n$$
where each $T_i$ consists of $m$ points: $$T_i = (p_1^i, p_2^i, \ldots, p_m^i)\}$$ 
for $1 \leq i \leq n$, and $p_j^i (j \in \{1,2,\ldots, m\})$ is a spatio-temporal point consisting of latitude and longitude values. Find a set of:
$$S_1, S_2, \ldots, S_n$$
such that we maximize the similarity between the two sets:

$$\max \text{ Similarity}(T,S) $$
where $T =\{T_1, T_2, \ldots, T_n\}, S =\{S_1, S_2, \ldots, S_n\}$, and similarity encompasses Path and Cluster similarity. For the sake of simplicity, we assume the number of synthetic trajectories is equal to the number of real trajectories. We discuss the metrics used in Section 3.5.

%[high dimensionality of embeddings vs small dataset]
\section{Related Works}
Trajectory generation in the wildlife domain is dominated by models based on Correlated Random Walk (CRW) \cite{bergman2000caribou}, Brownian motion \cite{horne2007analyzing}, and Levy Walk/Flight \cite{levy}. These methods are based on random steps and turning angles that follow a pre-defined distribution of step length. \textit{Trajr} is a popular wildlife trajectory R library that provides tools to generate trajectories using each of the aforementioned models \cite{trajr}. 

In \cite{technitis_b_2015}, the authors proposed a CRW trajectory generation method based on the maximum speed and expected direction of movement. However, the approach requires known starting and endpoints and is based on random distributions of velocity and movement angle. Nonetheless, such methods are more effective for simpler concerns such as missing data imputation, i.e. the estimation of missing points between two known points. They are more useful in capturing the micro movement of animals within their home range \cite{reynolds_scale-free_2009}.

On the other hand, the generation of trajectory in the pedestrian literature is constrained by road, other agents, and physical constraints. In \cite{trajvae}, an additional embedding layer is introduced to the VAE Encoder to account for the road network. This ensures the road features are embedded in the learning and understood by the generated trajectories. This has limited applicability in wildlife movement where constraints are terrain-based. On the other hand, a minimum bounding region (MBR) is capable of establishing constraints on the spatial areas that a trajectory may be allowed to navigate.

Generative Adversarial Networks (GANs) are another class of generative algorithms that have been utilized for the trajectory generation problem \cite{wang2021large}. However, GANs are well known for their poor convergence properties \cite{zhang2018convergence}. In addition, GANs generally require a large sample of training data and are thus ineffective for a sparse sample set \cite{trajvae}.
 \begin{figure}
\includegraphics[width=0.4\textwidth]{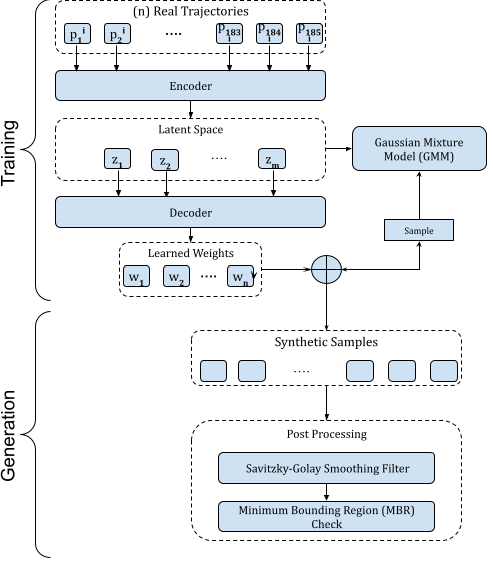} 
\centering
\caption{The proposed WildGEN Framework }
\begin{tabular}{r@{: }l r@{: }l} 
\end{tabular}
\label{fig:2}
\end{figure}
\section{Methodology}
Figure \ref{fig:2} depicts the architecture of the proposed WildGEN to generate and post-process a set of real trajectories. The framework is composed of four main components:
\begin{itemize}
    \item A Variational Auto Encoder (VAE): takes a data set consisting of a training set of n-day long trajectories, each of which represents a single latitude/longitude pair per day for each subject. 
    \item A Gaussian Mixture Model (GMM): learns the probability distribution of the training set in the latent space.
    \item A trajectory smoothing filter: utilizes Savitzky-Golay smoothing to reduce the excessive wandering effect observed on the generated trajectories.
    \item A Minimum Bounding Region (MBR): analyzes the generated trajectories violation of the constraints on acceptable movement region, and excludes outliers.
    
\end{itemize}

The framework consists of a training phase where the embeddings of each trajectory are learned in the latent space. This allows the GMM to learn the distribution of the training set in the latent space, and hence generate new trajectories based on the distribution. The trajectories are decoded using the learned weights from the decoding space. A smoothing post-processing step is utilized on each synthetic trajectory, followed by an MBR step that accepts or rejects it.

\subsection{Variational Autoencoders }

Variational Autoencoders (VAEs) \cite{kingma2019introduction} are a class of generative models that have gained significant traction in the field of trajectory generation due to their ability to capture complex data distributions and generate new samples. VAEs consist of two primary components: the Encoder and the Decoder.

\subsubsection{Encoder}

The Encoder, denoted as \(q(z | T)\), takes an input trajectory \(T\) and maps it to a latent space representation \(z\). The Encoder learns to capture the essential features and patterns present in the trajectory data. It compresses the high-dimensional trajectory space into a lower-dimensional latent space, effectively creating a data-driven embedding of the input trajectory.

\subsubsection{Decoder}

The Decoder, denoted as \(p(T | z)\), takes a point \(z\) from the latent space and generates a trajectory \(T\) in the original data space. The Decoder's role is to reconstruct trajectories from the latent space representations, effectively learning the weights that convert latent data back to the original data space.

\begin{figure*}[!htb]
  \subcaptionbox*{Real (green) vs Generated (blue)}[.33\textwidth]{%
    \includegraphics[width=\linewidth]{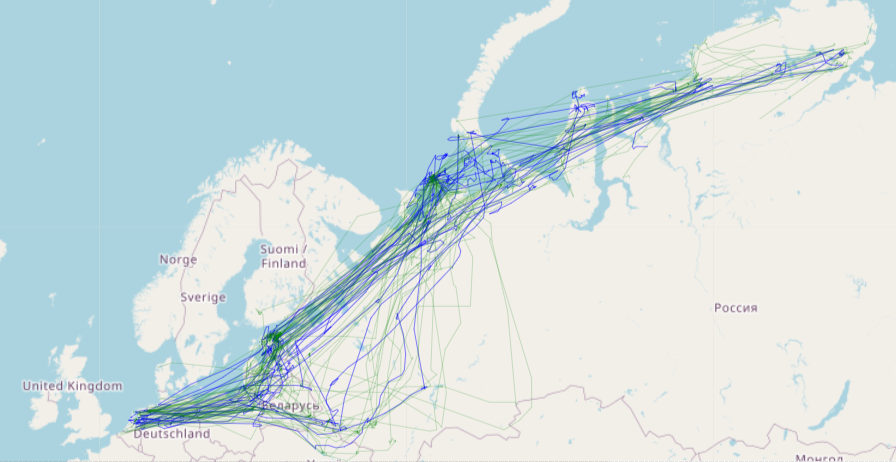}%
  }%
  \hfill
  \subcaptionbox*{Real (green) vs Generated (blue)}[.33\textwidth]{%
    \includegraphics[width=\linewidth]{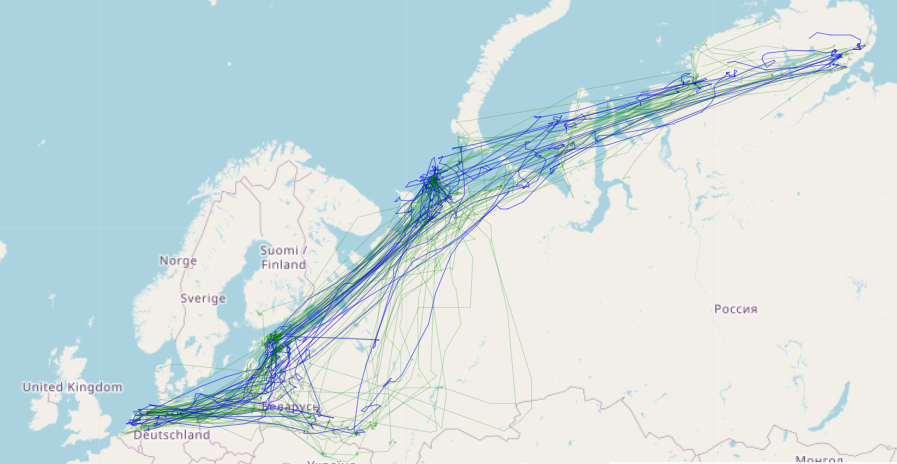}%
  }%
    \hfill
  \subcaptionbox*{Real (green) vs Generated (blue)}[.33\textwidth]{%
    \includegraphics[width=\linewidth]{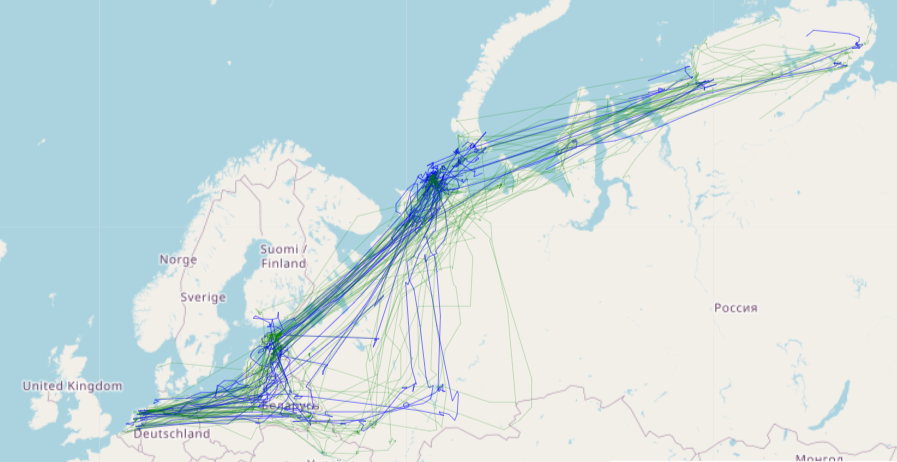}%
  }%
  \caption{
  Three different sets of Generated Trajectories mapped against a background of Real Trajectories
}
\end{figure*}

\subsection{Savitzky-Golay Smoothing}
Trajectory data, often obtained from GPS or motion tracking systems, can be inherently noisy due to measurement errors and environmental factors. Savitzky-Golay smoothing is a valuable tool for enhancing trajectory data quality while preserving essential trajectory characteristics \cite{schafer2011savitzky}.

By applying Savitzky-Golay smoothing to trajectory data, one can effectively reduce the noise present in position ($x_i$) or velocity ($v_i$) measurements. This noise reduction is achieved through a convolution operation using a set of smoothing coefficients ($c_j$) within a local window:

\[ y_i = \sum_{j=-n}^{n} c_j x_{i+j} \]

where $y_i$ represents the smoothed trajectory data at point $i$, $x_{i+j}$ are the neighboring data points within a window of size $2n+1$, and $c_j$ are the coefficients of the polynomial of order $m$ that minimize the mean squared error between the smoothed data and the noisy trajectory data.

The choice of polynomial order ($m$) and the window size ($2n+1$) in the context of trajectory data should be carefully considered. Too aggressive smoothing (high $m$) may lead to the loss of important trajectory details, while too little smoothing (low $m$) may not adequately mitigate noise. Savitzky-Golay smoothing, when appropriately tuned, is a valuable mathematical tool for improving the interpretability and reliability of trajectory data.

\subsection{Minimum Bounding Region}
Minimum Bounding Rectangles are the smallest axis-aligned rectangles (in two dimensions) or hyperrectangles (in higher dimensions) that enclose a set of trajectory points. In this paper, we generalize them using a convex hull and refer to them as Minimum Bounding Regions (MBR). By encapsulating trajectories within MBRs, we establish constraints on movement, such as maximum and minimum bounds in both spatial dimensions. 

The advantages of using MBRs lie in their simplicity and efficiency. They provide a compact representation of the potential spatial footprint of trajectory data, making it easier to detect outliers or unusual behavior. Moreover, MBRs facilitate quick assessments of whether trajectories remain within predefined geographical bounds, aiding in the enforcement of constraints in the environment, whether it is regional bounds, rivers, mountain ranges, and other physical constraints.

\subsection{Baselines}
In order to assess the performance of our framework we compare the results obtained to existing methods for animal trajectory generation. As discussed earlier, CRW-based and Brownian motion methods fail at generating long-horizon trajectories for tasks such as migration. Similarly, deep learning methods require large training samples. 

As such we utilize two baselines that appear to be plausible for the task of long-horizon trajectory generation, namely:

\begin{itemize}
    \item Levy Walk/Flight
    \item Heteroscedastic Multi-variate Gaussian Process Regression (Heteroscedastic GPR)
\end{itemize}

Next, we discuss the details of each of these methods, accordingly.

\subsubsection{Levy Walk/Flight} is a stochastic process that models movement using a Cauchy distribution to generate trajectory points, which are determined by a combination of step length, variance, and angular standard deviation. These parameters are derived from empirical data or real-world samples and are crucial inputs for the Levy trajectory generation process.

The algorithm combines these parameter values of step length, variance, and angular standard deviation with actual data points to produce the Levy trajectory, which is a dynamic and realistic path that captures the distinctive properties of the moving animal. 

\subsubsection{Heteroscedastic GPR}

Heteroscedastic GPR \cite{le2005heteroscedastic} is an extension of traditional Gaussian Process Regression (GPR) that explicitly models and accounts for varying levels of noise in the data. GPR is a powerful and versatile non-parametric method commonly used for modeling complex relationships in data. Traditionally, GPR assumes that the noise in the observed data is homoscedastic, meaning that the variance of the noise is constant across all data points. However, real-world data often exhibits heteroscedasticity, where the variance of the noise varies across the input space or with respect to the magnitude of the predictions. In such cases, traditional GPR may provide suboptimal results, leading to biased predictions and inaccurate uncertainty estimates.

Heteroscedastic GPR addresses the issue of varying noise levels by modeling the noise variance as a function of input variables. This is achieved through the introduction of a separate Gaussian process, often referred to as the noise process or the noise model. The noise process provides an estimate of the noise variance associated with each data point, which is then used to adjust the likelihood function in the GPR model. 

Given a dataset of $N$ observations $\{(x_i, y_i)\}_{i=1}^N$, where $x_i$ represents the input variable and $y_i$ represents the observed output variable.
The observed data $y_i$ is affected by heteroscedastic noise, meaning that the variance of the noise $\epsilon_i$ depends on the input variable $x_i$.

The model can be characterized by:
\begin{itemize}
    \item \textbf{Latent Function}: The underlying latent function $f(x)$ is modeled as a Gaussian process with a mean function $\mu(x)$ and a covariance (kernel) function $k(x, x')$, as in traditional GPR.
    \item \textbf{Noise Model}: A separate Gaussian process is introduced to model the noise variance as a function of the input variables. This is represented as $g(x)$, where $g(x_i)$ models the noise variance for each data point $x_i$.
\end{itemize}

The likelihood function in Heteroscedastic GPR is modified to account for the varying noise levels:

\[
p(y_i | f(x_i), g(x_i)) = \frac{1}{\sqrt{2\pi g(x_i)}} \exp\left(-\frac{(y_i - f(x_i))^2}{2g(x_i)}\right)
\]

Here:
\begin{itemize}
    \item $p(y_i | f(x_i), g(x_i))$ is the likelihood of observing $y_i$ given the latent function $f(x_i)$ and the noise variance $g(x_i)$.
    \item $g(x_i)$ represents the estimated noise variance for data point $x_i$.
    \item $f(x_i)$ is the value of the latent function at $x_i$.
    \item $y_i$ is the observed data point at $x_i$.
\end{itemize}

The need for Heteroscedastic GPR arises from the requirement to model multiple data points for a given input, where the input ($x_i$) is the day number in the trajectory, and the output ($y_i$) is the set of all real points from that day.
\subsection{Evaluation Metrics}
We base our selection of evaluation based on the objectives of trajectory generation, which we reiterate here for convenience:

\begin{enumerate}
    \item (Path Similarity) How similar are the points of the synthetic trajectories to the real trajectories?
    \item (Cluster Similarity) How well do the synthetic trajectories align with the real trajectories in terms of areas of concentration?
\end{enumerate}

In order to measure the first point, we use Hausdorff Distance similar to \cite{trajvae}, and we utilize the Pearson Correlation Coefficient to measure the similarity of the cluster sets for the second metric. Next, we describe both of them in further detail.

\subsubsection{Hausdorff Distance}
Hausdorff Distance is a valuable metric for assessing the similarity or dissimilarity between movement paths. Trajectory data can be compared using the Hausdorff distance to understand how closely two trajectories align or diverge.
The Hausdorff distance thus provides a way to measure the dissimilarity between two trajectories by considering the maximum minimum distance between points of one trajectory to the other trajectory.

The mathematical formulation of the Hausdorff distance (Hausdorff metric) between two sets of points, A and B, in a metric space can be expressed as follows:

Given two sets, A and B, each containing a number of points the Hausdorff distance from set A to set B, denoted as H(A, B), is defined as:

Hausdorff distance from set \(A\) to set \(B\):
\[H(A, B) = \max_{a \in A} \left( \min_{b \in B} d(a, b) \right)\]

Hausdorff distance from set \(B\) to set \(A\):
\[H(B, A) = \max_{b \in B} \left( \min_{a \in A} d(b, a) \right)\]

where d(a,b) is the distance function that computes the distance between two points.

\subsubsection{Coefficient of Correlation}
Pearson's coefficient of correlation \cite{asuero2006correlation}, denoted as "r" is a fundamental statistical measure used to quantify the strength and direction of linear relationships between two distributions. It is a widely used statistical measure that quantifies the strength and direction of a linear relationship between two distributions. Here are the key characteristics and aspects of Pearson's correlation coefficient:

\begin{itemize}
  \item An ``$r$'' value of -1 indicates a perfect negative linear relationship, where one variable decreases as the other increases in a perfectly linear fashion.
  \item An ``$r$'' value of 1 signifies a perfect positive linear relationship, where both variables increase together linearly.
  \item An ``$r$'' value of 0 implies no linear relationship, indicating that the variables are independent of each other.
\end{itemize}

The calculation of Pearson's correlation coefficient is based on the formula:

\[
r = \frac{\sum{(X_i - \bar{X})(Y_i - \bar{Y})}}{\sqrt{\sum{(X_i - \bar{X})^2}\sum{(Y_i - \bar{Y})^2}}}
\]

Here, $X_i$ and $Y_i$ represent individual data points from the two variables, and $\bar{X}$ and $\bar{Y}$ are their respective means.

\section{Experiments}
\subsection{Datasets}
The dataset used for our validation is The Western Palearctic greater white-fronted geese dataset \cite{dataset}. The dataset is hosted and was obtained from Movebank\footnote{https://www.movebank.org}. The dataset contains reading of 91 different birds over a period of several months to multiple years. We preprocessed the dataset using Moveapps \cite{kolzsch2022moveapps} to limit the observations to one observation per day. Next, the time frames were clipped to a window between March 1 to September 1 (185 days) to capture the northward migration. Any subjects that had significant readings missing during this time frame were dropped. Subjects that were tracked over multiple years were split into multiple samples. Simple data interpolation was utilized to fill in any remaining isolated gaps in the trajectories. The resultant dataset consists of 60 trajectories over a period of 185 days.

\subsection{Experimental Settings}
Here we discuss the implementation details of the WildGEN framework as well as the two baselines we compare it against.
 \begin{figure}
\includegraphics[width=0.5\textwidth]{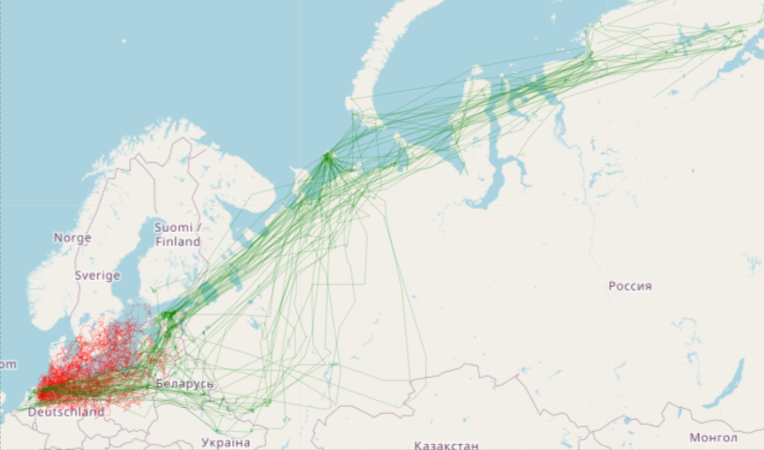} 
\centering
\caption{Real (green) vs Generated (red) Trajectories using Levy Walk/Flight}
%%\begin{tabular}{r@{: }l r@{: }l}
%%Green & Real Trajectories & Red & Generated Trajectories 
%%\end{tabular}
\label{fig:levy}
\end{figure}

\subsubsection{Levy Walk/Flight Implementation Details}

 \begin{figure}
\includegraphics[width=0.4\textwidth]{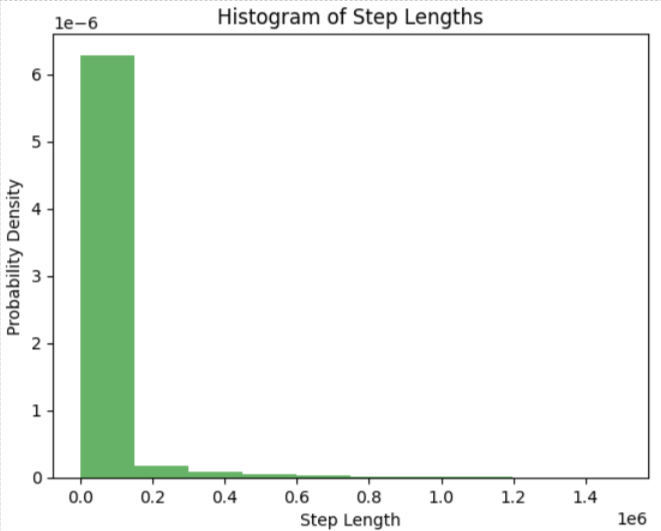} 
\centering
\caption{A Histogram of the step length distribution. Note that both axis have a power multiple}
%%\begin{tabular}{r@{: }l r@{: }l}
%%Green & Real Trajectories & Red & Generated Trajectories 
%%\end{tabular}
\label{fig:levyhis}
\end{figure}
A simple way of implementing Levy Walk/Flight involves analyzing the step lengths of the trajectory and learning the coefficient of Cauchy distribution ($\alpha$) from that data. However, the $\alpha$ value computed for the dataset was less than 1, which suggests that your step lengths have a very heavy-tailed distribution with an increased likelihood of extremely long steps as depicted by Figure \ref{fig:levyhis}. Nonetheless, for the purposes of this work, we used the Trajr R library \cite{trajr}, fed with the following input variables: 
\begin{itemize}
    \item Step Length
    \item Angular Error
    \item Linear Error
\end{itemize}

These variables were measured based on the training dataset. 

As per the documentation of Trajr \cite{trajr}, the generated trajectory is computed with direction = 0. In order to guide the movement in the correct northeast direction, the azimuth angle between the average starting point and the average ending point was computed. The generated trajectories were rotated counterclockwise with the computed angle.

Similar to WildGEN, the post-processing steps including Smoothing and MBR inclusiveness check were applied to the generated trajectories, and trajectories that violated the MBR were discarded.

Figure \ref{fig:levy} demonstrates the generated trajectories from Levy Walk/Flight.
 \begin{figure}
\includegraphics[width=0.5\textwidth]{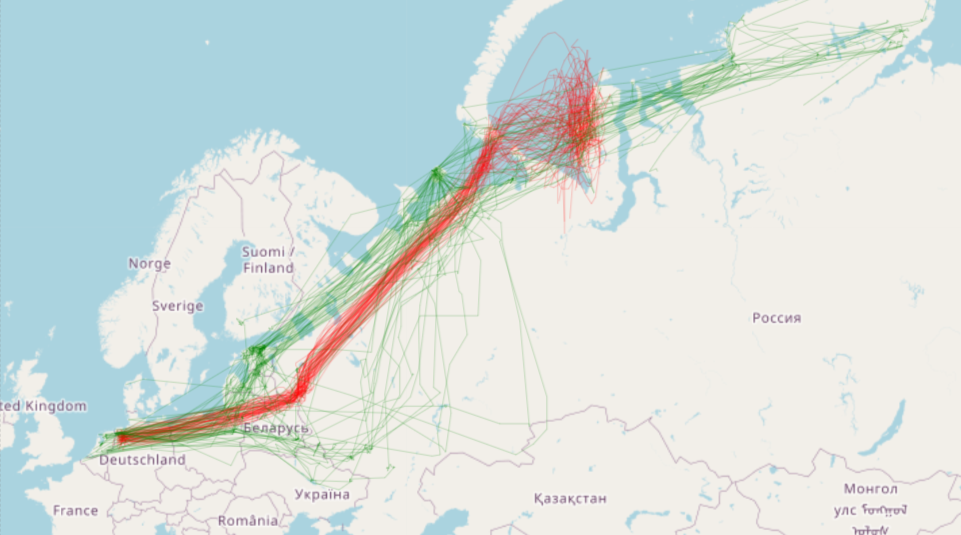} 
\centering
\caption{Real (green) vs Generated  (red) Trajectories using Heteroscedastic GPR}
%\begin{tabular}{r@{: }l r@{: }l}
%Green & Real Trajectories & Red & Generated Trajectories
%\end{tabular}
\label{fig:hgpr}
\end{figure}

\subsubsection{Heteroscedastic GPR Implementation Details}
The Python GPy\footnote{https://github.com/SheffieldML/GPy} library was used to model Heteroscedastic GPR using the $GPHeteroscedasticRegression$ class with a trained MLP and Bias kernels optimized on the training data. The GPR utilized a random set of 50\% of the training data. This reduced the computational requirements needed to train the Heteroscedastic GPR. 

In addition, the variances were computed using the norm of the $lng$ and $lat$ at each point. This helped limit overfitting of the model.

It is noteworthy that we also attempted to use the mean trajectory to model the GPR and train a Matern kernel based on the mean squared error against each training data sample. This approach was overfitting, and the samples generated were almost indistinguishable. 

Figure \ref{fig:hgpr} depicts trajectories generated by Heteroscedastic GPR compared to real trajectories. It can be observed that the clusters formed do not visually appear to align with the clusters of the real trajectories. 
\begin{figure}
\includegraphics[width=0.9\linewidth]{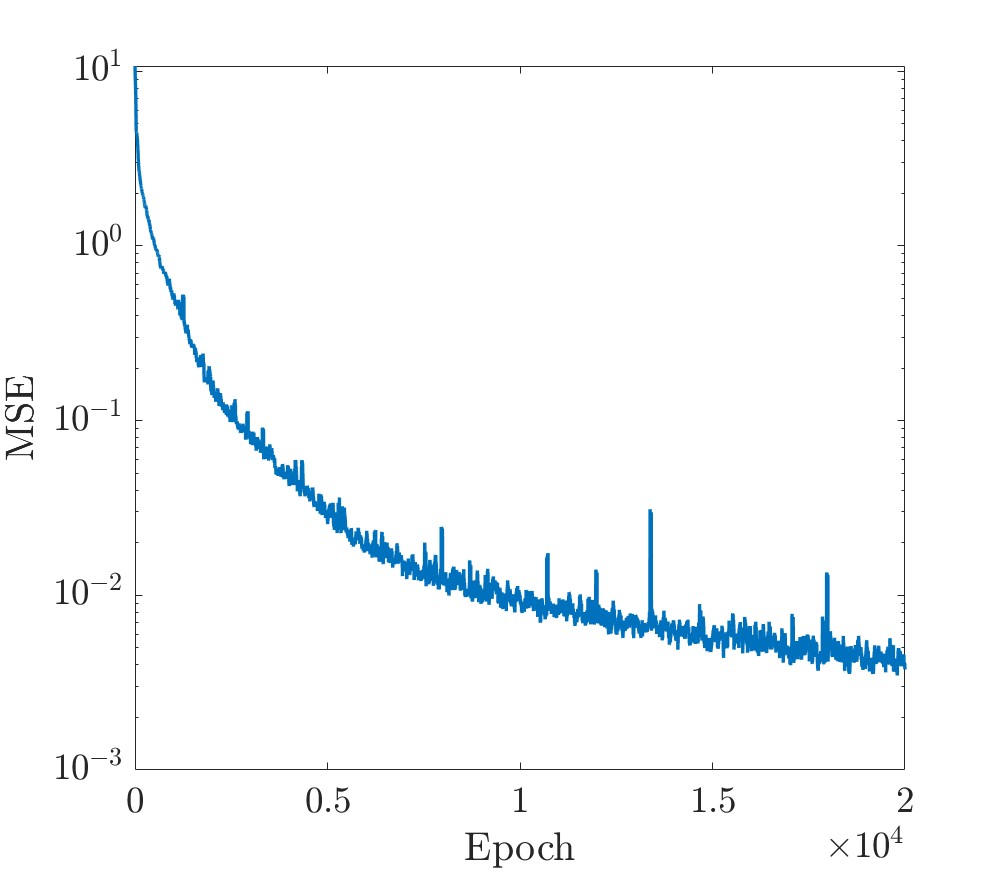} 
\centering
\caption{Training MSE}
%\begin{tabular}%{r@{: }l r@{: }l} 
%\end{tabular}
\label{fig:MSE}
\end{figure}

\begin{figure}
\includegraphics[width=0.9\linewidth]{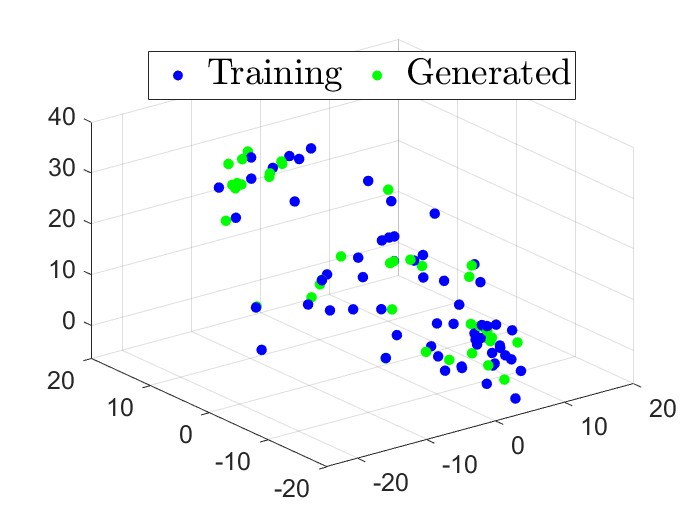} 
\centering
\caption{Compressed Space}
%\begin{tabular}%{r@{: }l r@{: }l} 
%\end{tabular}
\label{fig:latent}
\end{figure}

\subsubsection{WildGEN Implementation Details}
In the context of this study, the input structure comprises a dataset consisting of 60 records, each encompassing longitude and latitude coordinates, spanning a period of 185 days. These spatial-temporal records are meticulously rearranged into a vector representation composed of 370 inputs. It is noteworthy that prior to further processing, the input data undergoes a normalization step, which involves dividing each input value by 0.3 times the maximum input value, ensuring that the data adheres to a standardized scale.

The neural network architecture takes the form of an VAE. It initiates with an input layer of 370 neurons, which serves as the point of entry for the data. Sequentially, the network progresses through a hidden layer featuring 300 neurons, followed by another hidden layer consisting of 100 neurons. Subsequently, the architecture transitions into a lower-dimensional latent space representation, characterized by a dimensionality of 3. Following the latent space encoding, the network proceeds with two additional hidden layers, each comprising 100 neurons, facilitating the extraction of essential data features. The network culminates in a final hidden layer containing 300 neurons, serving as the penultimate stage. Ultimately, the network's journey concludes with an output layer designed to match the initial input's dimensionality, comprising 370 units. This VAE architecture effectively captures data patterns, allowing for both data compression and reconstruction. 

Activation functions are carefully chosen for each layer to regulate the flow of information through the neural network. The first hidden layer is governed by a modified Rectified Linear Unit (ReLU) activation function, while the second hidden layer employs a linear activation function. The latent space is also characterized by a linear activation function. Notably, the fourth and fifth hidden layers utilize a modified ReLU activation function, which is characterized in this work by a slope of 0.06 for positive values of the argument and a slope of 0.001 for negative values.

In this VAE-based study, the Mean Squared Error (MSE) is the primary loss function, assessing data reconstruction quality. The MSE, as shown in Figure\ref{fig:MSE}, quantifies the model's ability to successfully replicate input data within its latent space. Additionally, a Gaussian Mixture Model (GMM) with 15 components is fitted to the latent space using the 60 compressed samples shown in Figure \ref{fig:latent}, for deeper insight into data distribution. To generate trajectories, we employ a sampling approach within the latent space, drawing samples from the GMM, as showcased in Figure \ref{fig:latent}, and subsequently propagate these samples through the decoder network.

\begin{figure}
\includegraphics[width=0.45\textwidth]{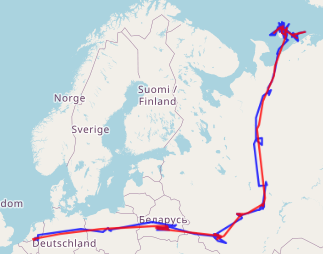} 
\centering
\caption{Savitzky-Golay Smoothing Filter: Before (blue) and After (red) for a sample generated trajectory}

\label{fig:4}
\end{figure}

\textbf{Smoothing}:
The Savitzky-Golay Smoothing was implemented against each trajectory based on a manual visual tuning of the parameters, namely window size and polynomial order. As Fig \ref{fig:4} demonstrates, the filter can reduce excessive wandering of the generated trajectories in comparison with the real trajectories.

\textbf{MBR}:
The Minimum Bounding Region (MBR) was calculated from the real trajectories to constrain the region in which the birds can fly. In different problems, one or multiple MBRs can be used to signify water bodies, impenetrable mountain ranges, and other terrain-based obstacles. The MBR step discards any trajectory that does not fully lie inside the calculated MBR. This has resulted in discarding 29.6\% of the generated trajectories and retaining the remaining trajectories.

 \begin{figure}
\includegraphics[width=0.45\textwidth]{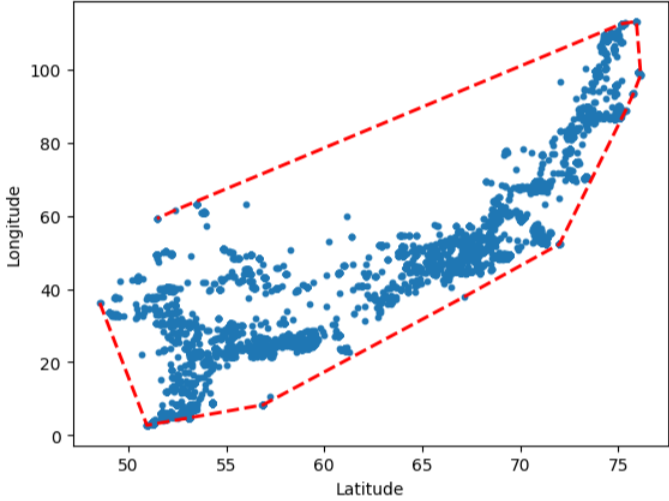} 
\centering
\caption{A Minimum Bounding Region based on the real trajectories}
\begin{tabular}{r@{: }l r@{: }l} 
\end{tabular}
\label{fig:mbr}
\end{figure}

\subsection{Model Comparison}
\begin{figure}[!tbp]
  \centering
  \subfloat[Sillouhette score ]{\includegraphics[width=0.49\linewidth]{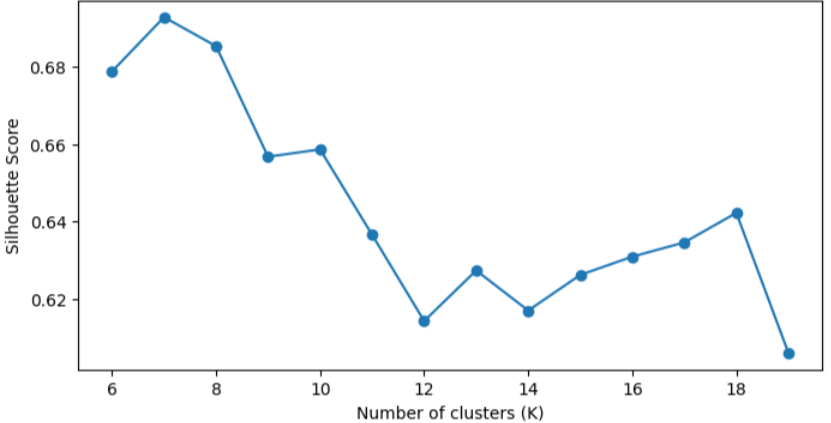}\label{fig:sill}}
  \hfill
  \subfloat[Distortion (Elbow Method)]{\includegraphics[width=0.49\linewidth]{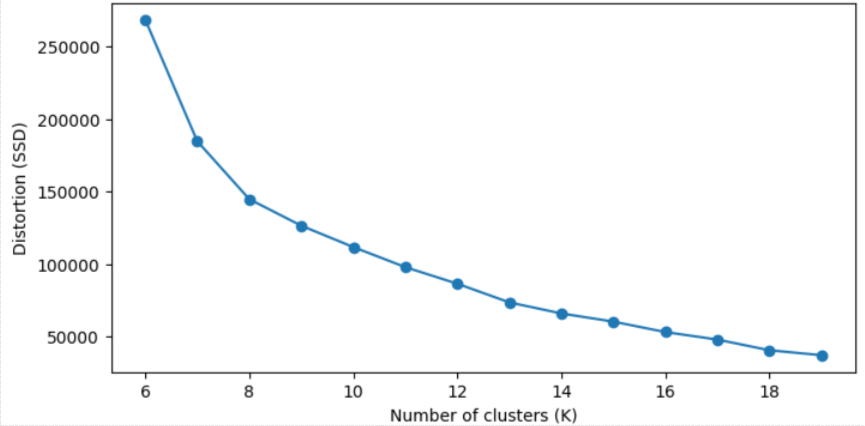}\label{fig:elbo}}
  \caption{Choosing the best value of K for K-means clustering based on two separate methods}
\end{figure}

In order to compare the models, the first 60 eligible trajectories from each model's synthetic set were used to match the set of real trajectories on the evaluation metrics. The remaining trajectories were not used during the benchmarking process in this step. 

\subsubsection{Path Similarity}
The Hausdorff distance between each pair of trajectories from the synthetic and real set were compared. Only the lowest value was retained for each synthetic trajectory. In other words, for each generated trajectory, the distance was computed with the trajectory closest to it from the real trajectory set. 

We report the minimum, mean, and average distance for each model in Table \ref{tab:example}.

\subsubsection{Cluster Similarity Calculation}
K-means clustering was performed on the set of all points, i.e. 

$$\bigcup_{i} \{ p_1^i, p_2^i, \ldots, p_n^i \}$$ for all $T_i$. The value of $K$ was chosen using a Silhouette Score and validated visually using the Elbow method \cite{saputra2020effect}. The number of clusters based on the geographic projection of the real trajectory should be between 10 and 15; K=13 was used in our experiments.

The trained clustering model was applied to the generated trajectory points of each model, and the Pearson Correlation Coefficient was calculated based on the distribution of points in each cluster.

\subsubsection{Results}
From Table \ref{tab:example}, we can observe that WildGEN provides a better overall Hausdorff Distance on average than the other models. Further, the clusters of WildGEN have better correspondence to the real trajectories. While Heteroscedastic GPR performs well on the Hausdorff distance metric, the results are influenced by the low variance in the model. It is visually apparent that trajectories generated by Heteroscedastic GPR follow the same path with little variance through the first half of the trajectory as depicted in Figure \ref{fig:hgpr}.

On the other hand, it can be observed that Levy Walk/Flight fails to generate any realistic trajectories. The metrics also indicate that the trajectories generated fail to capture any properties of the real trajectories. This is not surprising given the limited amount of information about the real trajectories that can be fed into the algorithm.

However, WildGEN performs well on the metrics in line with visual inspection. It performs significantly better than Levy Walk/Flight on average Hausdorff Distance and provides a 15.5\% improvement on average compared to Heteroscedastic GPR. There is a very high correlation between the cluster distributions and the generated points are closer on average to real trajectory points which is almost double the cluster similarity observed using Heteroscedastic GPR.

\begin{table}[H]
   \caption{Test Results for Hausdorff Distance and Pearson Correlation Coefficient} 
   \label{tab:example}
   \small
   \centering
   \begin{tabular}{l||ccc|c}
   \toprule\toprule
   \multirow{2}{*}{\textbf{Method}}  & \multicolumn{3}{c|}{\textbf{Hausdorff Distance}} & \textbf{Pearson (r)} \\ 
   &Min       &Max &Avg\\
   \midrule
   Levy Walk/Flight  & 22.251 & 39.799 & 30.590 & -0.0669\\
   Heteroscedastic GPR & 5.760 & \underline{7.155} & 6.208 & 0.4563\\
   WildGEN & \underline{3.542} & 7.450 & \underline{5.244} & \underline{0.9936}\\
   \bottomrule
   \end{tabular}
\end{table}

\begin{figure*}
  \subcaptionbox*{(a) Raw }[.24\textwidth]{%
    \includegraphics[width=\linewidth]{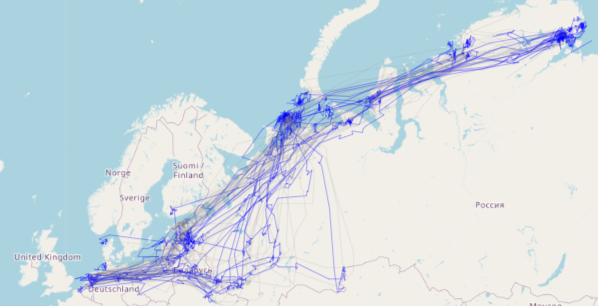}%
  }%
  \hfill
  \subcaptionbox*{(b) Smoothed }[.24\textwidth]{%
    \includegraphics[width=\linewidth]{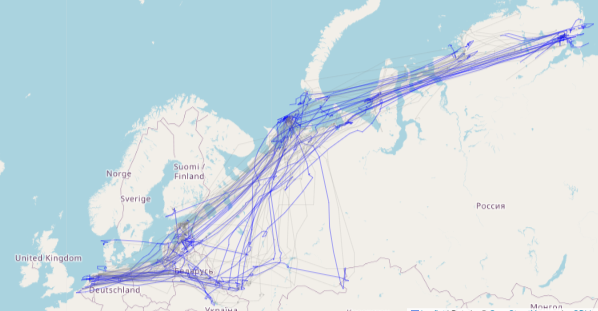}%
  }%
    \hfill
  \subcaptionbox*{(c) MBR Filtered}[.24\textwidth]{%
    \includegraphics[width=\linewidth]{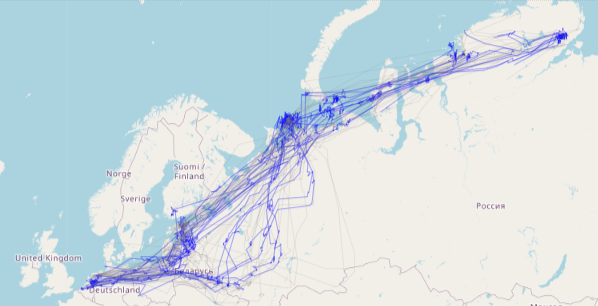}%
  }%
      \hfill
  \subcaptionbox*{(d) WildGEN}[.24\textwidth]{%
    \includegraphics[width=\linewidth]{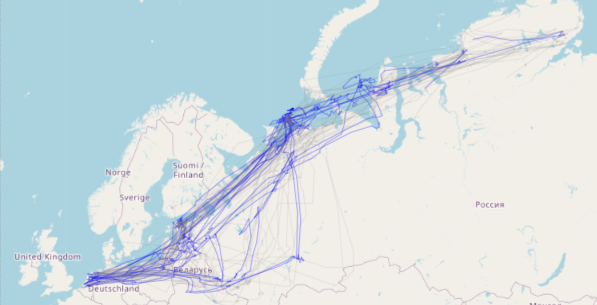}%
  }%
  \caption{The effect of different components on WildGEN on the generated trajectories}
\end{figure*}

\subsection{Ablation Study}
In the ablation study, we want to measure the effects of Smoothing and Minimum Bounding Region (MBR) on the metrics measured. 

\begin{table}[H]
   \caption{Test Results for Hausdorff Distance and Pearson Correlation Coefficient} 
   \label{tab:example}
   \small
   \centering
   \begin{tabular}{l||ccc|c}
   \toprule\toprule
   \multirow{2}{*}{\textbf{Method}}  & \multicolumn{3}{c|}{\textbf{Hausdorff Distance}} & \textbf{Pearson (r)} \\ 
   &Min       &Max &Avg\\
   \midrule
   No Post Processing  & 0.306 & 9.440 & 4.172 & 0.9881\\
   Smoothing Only & 3.542 & 10.047 & 5.565 & 0.9865\\
   MBR Only & \underline{0.395} & \underline{6.425} & \underline{3.849} & 0.9928\\
   WildGEN & 3.542 & 7.450 & 5.244 & \underline{0.9936}\\
   \bottomrule
   \end{tabular}
\end{table}

In order to perform the ablation study correctly, we ran the experiments on a random sample of trajectories, applied the specified method on the same set, and retained only the first 60 trajectories in line with the experiments in the previous section. 

The outcomes of our ablation study reveal that smoothing alone yields diminished results of the Hausdorff distance and Pearson Correlation Coefficient. However, as discussed previously, smoothing effectively addresses a specific visual challenge, notably the mitigation of excessive wandering. To provide a mathematical evaluation of the advantages provided by smoothing, we may need to consider additional metrics.

On introducing MBR to our trajectories, the measurable impact of smoothing diminishes. Nevertheless, the combined application of both smoothing and MBR yields a small improvement in the Pearson Correlation Coefficient. It can be observed that the penalization against the Hausdorff distance metric extends to the WildGEN scenario.

Our findings also serve to validate the viability of the raw trajectories generated by the VAE, as corroborated by the metrics introduced. Notably, the utilization of smoothing significantly contributes to their improved visual inspection when compared to real trajectory data.

It is important to note that comparing Smoothing Only and MBR Only scenarios is not completely objective. This is because MBR excludes many trajectories that the smoothing step retains. On the other hand, if we doctor the random set to ensure that only trajectories that meet the MBR criteria are used in the ablation study, it would defeat the purpose of the improvements of MBR. In that case, MBR and No Post Processing would have identical results.

\section{Conclusions and Future Work}
This work presented a novel method by which synthetic trajectories may be generated on a long horizon. The framework, WildGEN, can be characterized as follows:
\begin{itemize}
    \item Generate near-real, long-horizon trajectories based on a relatively small number of real trajectories

    \item Performs post-processing steps to further improve the generated trajectories 

    \item Shows improved accuracy compared to the state-of-the-art methods
\end{itemize}

Based on the findings from this work, there are multiple avenues for future work. A key limitation identified is the evaluation metrics used are not sufficiently representative of a \textit{good trajectory}. In particular, Hausdorff distance fails to capture the enhancement in the trajectory, and it is necessary to test other metrics such as Frechet Distance to better assess the effect of smoothing. It is also interesting to identify and utilize other potential smoothing filters.

In addition, as a follow-up to this work, we intend to:
\begin{itemize}
    \item Utilize other generative models, including normalizing flows that can provide the pertinent statistical weights of the generated trajectories 

    \item Extend and further investigate the applicability of WildGEN for various wildlife trajectory scenarios, including land-based and water-based animals.  
\end{itemize}


\bibliographystyle{ACM-Reference-Format}
\bibliography{sample-base}

\end{document}